# Fine-Tuning ChemBERTa for Predicting Inhibitory Activity Against TDP1 Using Deep Learning

**Baichuan ZENG**

Department of Computer Science and Engineering

The Chinese University of Hong Kong

baichuan@link.cuhk.edu.hk

## Abstract

Predicting the inhibitory potency of small molecules against Tyrosyl-DNA Phosphodiesterase 1 (TDP1) — a key target in overcoming cancer chemoresistance—remains a critical challenge in early drug discovery. We present a deep learning framework for the quantitative regression of pIC50 values from molecular Simplified Molecular Input Line Entry System (SMILES) strings using fine-tuned variants of ChemBERTa, a pre-trained chemical language model. Leveraging a large-scale consensus dataset of 177,092 compounds, we systematically evaluate two pre-training strategies—Masked Language Modeling (MLM) and Masked Token Regression (MTR)—under stratified data splits and sample weighting to address severe activity imbalance which only 2.1% are active. Our approach outperforms classical baselines Random Predictor in both regression accuracy and virtual screening utility, and has competitive performance compared to Random Forest, achieving high enrichment factor EF@1% 17.4 and precision Precision@1% 37.4 among top-ranked predictions. The resulting model, validated through rigorous ablation and hyperparameter studies, provides a robust, ready-to-deploy tool for prioritizing TDP1 inhibitors for experimental testing. By enabling accurate, 3D-structure-free pIC50 prediction directly from SMILES, this work demonstrates the transformative potential of chemical transformers in accelerating target-specific drug discovery. Code and models are publicly available at: https://github.com/NokkTsang/AIST4010_Project.

## 1 Introduction

The discovery of potent and selective inhibitors of Tyrosyl-DNA Phosphodiesterase 1 (TDP1) represents a promising strategy to overcome resistance in cancer chemotherapy. TDP1 is a key DNA repair enzyme that resolves topoisomerase I (Top1)-DNA covalent complexes; its overexpression is associated with resistance to Top1 poisons such as camptothecin and its derivatives. Consequently, pharmacological inhibition of TDP1 can potentiate the efficacy of existing chemotherapeutics, making it a compelling target for rational drug design.

Traditional approaches to TDP1 inhibitor discovery have relied heavily on labor-intensive high-throughput screening (HTS) or structure-activity relationship (SAR) studies based on small, congeneric compound series. While these methods have yielded valuable chemical probes, they are often limited by high costs, low throughput, and poor generalizability. In parallel, conventional machine learning models—such as random forests or support vector machines trained on hand-crafted molecular descriptors—have shown moderate success but struggle to capture the intricate, non-linear patterns embedded in molecular structure–activity relationships.

Recent advances in self-supervised representation learning offer a transformative alternative. Pre-trained chemical language models, such as ChemBERTa, leverage large-scale unlabeled molecular data, for example millions of Simplified Molecular Input Line Entry System (SMILES) strings, to learn contextualized representations that encode rich chemical and topological information. These models can be efficiently fine-tuned on modest labeled datasets for specific bioactivity prediction tasks, eliminating the need for manual feature engineering or 3D structural data.

In this work, ChemBERTa is fine-tuned—a RoBERTa-based transformer pre-trained on 77 million chemical structures—for the regression task of predicting pIC50 values against TDP1 using a large-scale dataset of 177,092 SMILES–pIC50 pairs. Our approach departs from prior studies by: focusing on continuous pIC50 regression rather than binary classification, preserving the full dynamic range of inhibitory potency; rigorously comparing two pre-training objectives—Masked Language Modeling (MLM) and Masked Token Regression (MTR)—to assess their impact on quantitative prediction; employing stratified data splitting and sample weighting to address severe activity imbalance; and benchmarking against the classical baseline like Random Forest to contextualize the added value of deep representation learning.

The resulting model will be provided in collaboration with members from the Faculty of Medicine at The Chinese University of Hong Kong to prioritize compounds for experimental validation in the Wet Lab, thereby establishing a closed-loop AI-to-wet-lab discovery pipeline. This demonstrates that fine-tuned chemical transformers can serve as powerful, general-purpose tools for target-specific inhibitor prediction, significantly accelerating early-stage drug discovery with minimal preprocessing and no reliance on 3D conformational data.

## 2 Related Work

The quest for Tyrosyl-DNA Phosphodiesterase 1 (TDP1) inhibitors has evolved from structure-based design to data-driven computational approaches, reflecting broader trends in cheminformatics and AI-enabled drug discovery. Early efforts focused on small, congeneric series, while recent work leverages large-scale bioactivity data and deep learning. Our study situates itself at the intersection of these paradigms, utilizing a consensus dataset of unprecedented scale for TDP1 and fine-tuning modern chemical language models for quantitative potency prediction.

### 2.1 Traditional QSAR and Structure-Based Approaches

Initial TDP1 inhibitor discovery relied on medicinal chemistry guided by structure–activity relationships (SAR) within narrow chemical series. Gladkova et al. reported the first berberine-derived TDP1 inhibitors, establishing a linear QSAR model with $R^2 = 0.65$ on a dataset of only 30 analogs, highlighting the limitations of small-data approaches in capturing complex SAR [1].

Similarly, Khomenko et al. explored usnic acid derivatives using molecular docking to prioritize compounds for synthesis, yielding dual TDP1/Top1 inhibitors but without quantitative activity modeling [2]. While valuable for lead optimization within a scaffold, such methods lack generalizability across diverse chemical spaces.

To address this, Kovaleva et al. integrated resin acid and adamantane moieties into novel TDP1 inhibitors and employed genetic algorithms to optimize multiple linear regression models, achieving $R^2 = 0.82$ on a modest dataset of 45 compounds [3]. Although more sophisticated, these descriptor-based models remain constrained by hand-crafted features and limited data volume.

## 2.2 Machine Learning on Large-Scale Bioactivity Data

A landmark resource in this context is the consensus bioactivity dataset introduced by Isigkeit et al. [4], which integrates bioactivity data from five major sources—ChEMBL, PubChem, BindingDB, IUPHAR/BPS Guide to Pharmacology, and Probes&Drugs—into a unified compendium of 1.14 million compounds and 10.9 million activity records across 5,613 targets. Through rigorous cross-source validation, this dataset achieves high chemical diversity and target coverage, making it particularly suitable for training robust machine learning models in chemogenomics.

Leveraging such resources, Lai et al. trained SMILES-based machine learning classifiers like Random Forest on TDP1 data from PubChem, achieving AUC = 0.85 for binary activity prediction [5]. However, their work treats inhibition as a classification problem, discarding the continuous potency information essential for lead optimization. Similarly, while high-quality quantitative screening data like $AC_{50}$ values for TDP1 are available in PubChem [6,7,8], most downstream analyses to date have focused on hit identification rather than regression modeling.

## 2.3 Deep Learning and Chemical Language Models

Recent advances in self-supervised learning have revolutionized molecular property prediction. ChemBERTa, introduced by Chithrananda et al. , adapts the RoBERTa architecture to SMILES strings, pre-training on 10 million unlabeled molecules via Masked Language Modeling (MLM) [9]. It achieves state-of-the-art results on benchmarks like solubility prediction $R^2 = 0.91$, demonstrating that transformers can learn rich chemical semantics from sequence alone.

Crucially, a variant—ChemBERTa-MTR—was pre-trained using Masked Token Regression (MTR), where the model predicts continuous molecular properties like logP and molecular weight of masked substructures [10]. This objective explicitly aligns with downstream regression tasks, making MTR a theoretically superior starting point for $pIC_{50}$ prediction compared to the reconstruction-focused MLM.

ChemBERTa's efficacy has been validated across targets. For instance, Huang et al. included TDP1 in the Therapeutics Data Commons (TDC) benchmark and demonstrated that transformer-based models outperform graph neural networks (GNNs) in several regression tasks [11]. However, no study to date has fine-tuned ChemBERTa specifically for TDP1 $pIC_{50}$ regression on a large, integrated dataset.

Building on this foundation, Ross et al. represents a significant scaling of the chemical language modeling paradigm [12]. Pretrained on 1.1 billion SMILES from PubChem and ZINC using a linear-attention Transformer with rotary positional embeddings, MoLFormer leverages masked language modeling at an unprecedented scale. Despite using only 2D SMILES as input, attention analysis reveals that MoLFormer implicitly learns 3D spatial relationships between atoms, enabling it to capture structural and electronic features critical for bioactivity. It achieves state-of-the-art performance across ten MoleculeNet benchmarks—including regression tasks like lipophilicity (logP) and solubility—outperforming both graph neural networks and earlier language models like ChemBERTa. Its success underscores the power of large-scale pretraining for learning transferable molecular representations directly from sequence, making it a strong candidate for $pIC_{50}$ prediction tasks such as ours. However, since the model is larger than ChemBERTa for fast processing, this option is abandoned.

## 2.4 Positioning of the Work

Our work advances TDP1 inhibitor prediction through a systematic comparison across three key dimensions. First, three PubChem assays and the consensus dataset Zenodo is integrated to construct the largest TDP1-specific resource to date with 177,092 compounds, formulated as a continuous $pIC_{50}$ regression task to support lead optimization. Second, two data handling strategies are benchmarked—stratified oversampling versus sample weighting—and find the latter better preserves activity distribution and improves early enrichment. Third, the impact of hyperparameter tuning with Optuna [13] and model architecture is evaluated, providing empirical evidence that ChemBERTa-MTR with sample weighting and optimized hyperparameters achieves superior virtual screening performance. By prioritizing drug discovery–oriented metrics like Enrichment Factor and Precision@K, a practical, reproducible framework is established that moves chemical foundation models beyond generic benchmarks toward target-specific quantitative prediction.

# 3 Data

Our study utilizes a large-scale, integrated dataset of TDP1 inhibitory activity compiled from three primary PubChem BioAssay records and one publicly available consensus bioactivity resource. The final curated dataset comprises 177,092 unique compounds, each represented by a canonical SMILES string and a continuous $pIC_{50}$ value, reflecting inhibitory potency against human TDP1.

## 3.1 Data Sources

The dataset is constructed from the following four sources: **PubChem BioAssay AID 485290** [6]. This quantitative high-throughput screening (qHTS) assay identifies inhibitors of TDP1 from a large chemical library. After rigorous filtering for confirmed "inhibitor" phenotype, valid SMILES, and positive $AC_{50}$ values 0.5 nM – 141.3 μM, 6,438 unique compounds were retained. **PubChem BioAssay AID 489007** [7]. A complementary screening campaign with a distinct compound library and assay protocol. This smaller but high-quality dataset contributed 664 unique inhibitors with $AC_{50}$ in range 4 nM – 158.5 μM. **PubChem BioAssay AID 686978** [8]. A large-scale screening effort yielding 172,535 unique compounds with $AC_{50}$ in range 0.8 nM – 82.8 μM. This assay forms the backbone of our dataset, providing broad chemical diversity. **Consensus Bioactivity Dataset Zenodo Record 6398019** [14]. A pre-processed, deduplicated collection of bioactivity data derived from a consensus of five major databases includes ChEMBL, PubChem, BindingDB, IUPHAR/BPS and

Probes&Drugs. From this resource, 454 TDP1-specific pIC$_{50}$ entries is incorporated that showed no overlap with the PubChem sources.

## 3.2 Data Processing and Curation

To ensure data integrity and consistency, a multi-stage curation pipeline is implemented: Duplicate Resolution: Cross-dataset duplicate SMILES were identified (2,808 total). For entries with identical SMILES but conflicting AC$_{50}$ values (2,790 cases), the arithmetic mean of the AC$_{50}$ values was computed before conversion to pIC$_{50}$. Unit Conversion: All AC$_{50}$ values (in μM) were converted to pIC$_{50}$ using the standard formula:

$$\mathrm{pIC}_{50} = -\log_{10}(\mathrm{AC}_{50} \times 10^{-6})$$

Quality Control: Entries were filtered to retain only those with non-null, valid SMILES strings and positive, numerical AC$_{50}$ values. The final dataset contains zero duplicates, missing values, or invalid structures.

## 3.3 Statistical Overview

The final curated dataset comprises 177,092 unique compounds, each annotated with a continuous pIC$_{50}$ value representing its inhibitory potency against TDP1. The pIC$_{50}$ values span a range of 3.85 to 9.10, with a mean of 4.80 ± 0.39 and a median of 4.69, indicating a right-skewed distribution—a characteristic commonly observed in high-throughput screening campaigns where the majority of screened compounds exhibit weak or negligible activity.

Using the conventional threshold of pIC$_{50}$ ≥ 6.0 to define activity, the dataset contains 3,800 active compounds (2.1%) and 173,292 inactive compounds (97.9%), resulting in a pronounced class imbalance with an inactive-to-active ratio of 45.6:1. Further stratification by potency reveals that only 792 compounds (0.4%) are highly potent (pIC$_{50}$ > 7.0, IC$_{50}$ < 100 nM), while the vast majority (148,415 compounds, 83.8%) fall into the weakly active range (pIC$_{50}$ 4.0–5.0). This heavy concentration of low-potency compounds presents a significant challenge for machine learning models, which may struggle to generalize to the sparse, high-value region of the activity landscape.

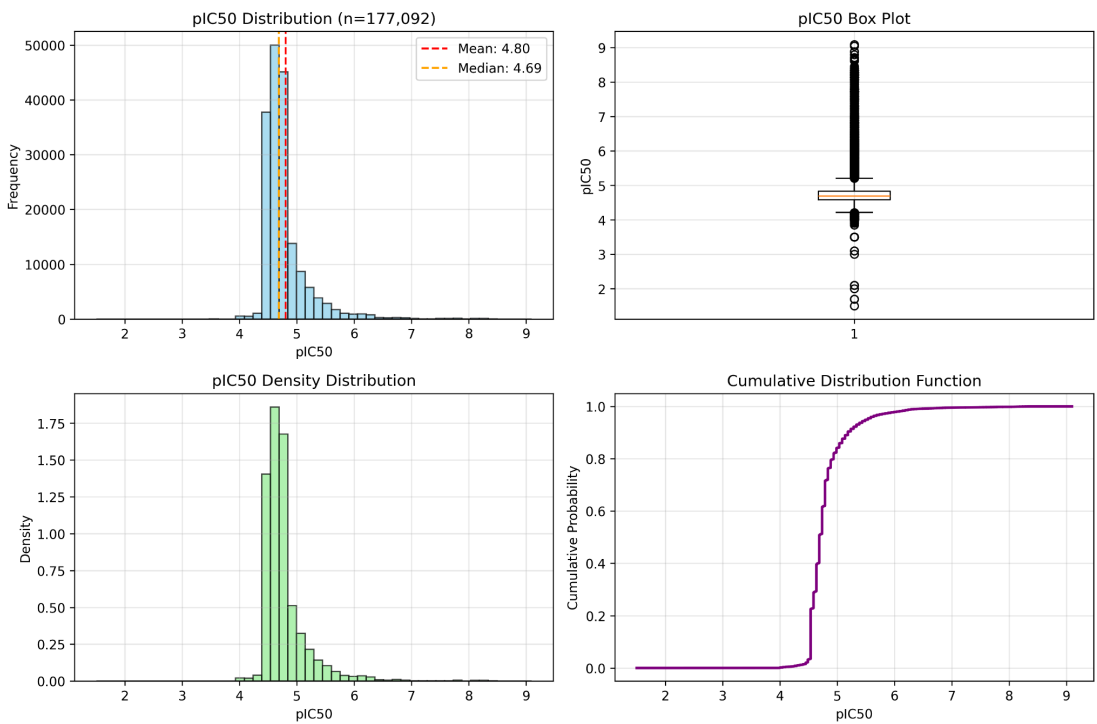

Figure 1: Data Analysis of the final TDP1 pIC$_{50}$ dataset

### 3.4 Data Format and Availability

The final curated dataset is distributed as a UTF-8–encoded comma-separated values (CSV) file comprising two columns: SMILES and pIC$_{50}$. A sample of the data is shown below:

Table 1: Sample entries from the final TDP1 pIC$_{50}$ dataset

| SMILES | pIC50 |
| --- | --- |
| CCC(C)(C#C)O | 9.096910013008056 |
| C1C=C[C@@H]2CC(=O)N(CC3=CC=CC1=C23)CC4=CC=C(C=C4)Br | 4.949999401714238 |

This rigorously processed dataset—free of duplicates, invalid structures, and missing values—constitutes a high-integrity resource for the development and evaluation of quantitative structure–activity relationship (QSAR) models. By preserving the continuous nature of pIC$_{50}$ and reflecting the true activity distribution of large-scale screening campaigns, it offers a realistic and challenging benchmark for modern machine learning approaches in TDP1 inhibitor discovery and, more broadly, in cheminformatics.

## 4 Approch

This section presents a systematic methodology to identify and validate the optimal fine-tuning strategy for ChemBERTa in the task of TDP1 pIC$_{50}$ regression from SMILES strings. Rather than adopting a single configuration, conducting a controlled ablation study across three critical dimensions to rigorously justify our final pipeline: data handling strategy, comparing stratified oversampling against sample weighting to address severe activity imbalance; hyperparameter optimization, evaluating models trained with fixed hyperparameters versus those tuned via Optuna Bayesian optimization; and model architecture, benchmarking ChemBERTa variants against classical machine learning baselines.

By holding the data split, random seed, and evaluation protocol constant across all experiments, we ensure a fair and reproducible comparison. This multi-axis analysis allows us to isolate the contribution of each design choice and empirically demonstrate that the combination of ChemBERTa-MTR, sample weighting, and Optuna-tuned hyperparameters yields the great effective model for virtual screening—an approach whose superiority is validated not by assumption, but by comprehensive experimental evidence.

## 4.1 Data Preprocessing and Stratified Splitting

Given the extreme activity imbalance in the dataset—where active compounds (defined as pIC$_{50}$ ≥ 6.0) constitute only 2.1% of the total 177,092 entries, yielding an inactive-to-active ratio of approximately 45.6:1—a stratified partitioning strategy is essential to ensure statistical reliability and reproducibility in model evaluation. The dataset is split into training 70%, validation 15%, and test 15% subsets using stratified sampling based on the binary activity label derived from the pIC$_{50}$ ≥ 6.0 threshold. This guarantees that each subset maintains the same activity prevalence around 2.1%, thereby mitigating sampling-induced variance in performance metrics across experimental runs. Two distinct training strategies are evaluated for deep learning models. The first, oversampling, balances the training set to a 50:50 active–inactive ratio by random sampling with replacement of active compounds, thereby creating an artificial but class-balanced distribution for initial benchmarking. The second, sample weighting, preserves the original training distribution and instead assigns a per-sample weight $w_i$ during loss computation to compensate for class imbalance. Specifically, the weight is defined as

$$w_i = \frac{N}{2N_{c_i}},$$

where $N$ is the total number of training samples and $N_{c_i}$ is the cardinality of the activity class to which the compound *i* belongs. Under this scheme, active compounds receive a weight of approximately 23.3, while inactive compounds receive a weight of 0.51, preserving the original 45.6:1 imbalance as an effective learning emphasis ratio. Crucially, validation and test sets remain in their native imbalanced form across all experiments to reflect real-world virtual screening conditions.

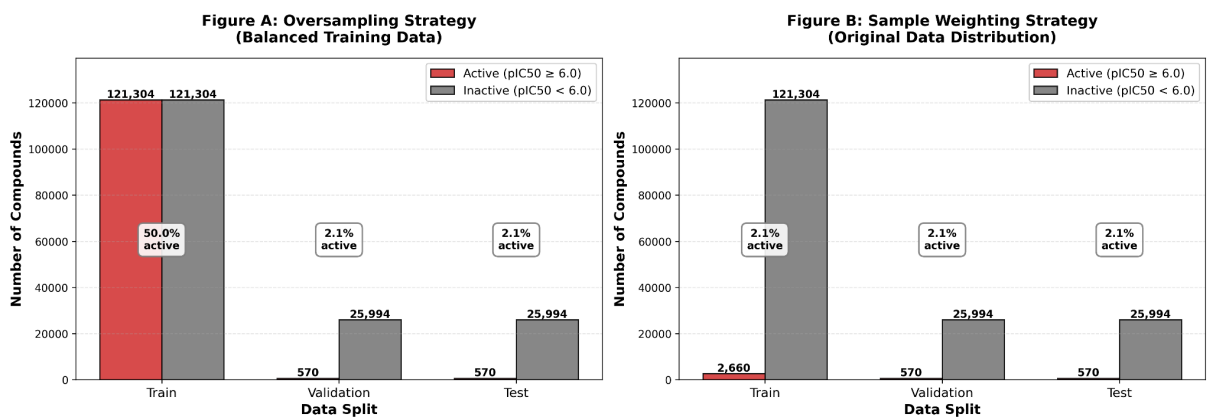

Figure 2: Data Distributions over two strategies

## 4.2 Baseline Models

Two baseline models are implemented to establish performance bounds and contextualize the added value of deep learning approaches. The first is a random prediction baseline, which generates $pIC_{50}$ predictions by sampling uniformly from the observed range of the training set. This provides a theoretical lower bound: any meaningful model must demonstrably outperform this null hypothesis in both regression and virtual screening metrics. The second baseline is a Random Forest regressor trained on a rich set of hand-crafted molecular features. All features are computed using RDKit, standardized via a StandardScaler, and subjected to basic quality control like removal of zero-variance features. The model is configured with 100 estimators, a maximum depth of 20, and a fixed random seed for reproducibility, and is trained on the same stratified splits used for all other models. Its strong performance in prior cheminformatics studies makes it a stringent benchmark for evaluating the necessity of end-to-end deep learning.

## 4.3 Deep Learning Model: ChemBERTa Architectures

Two variants of the ChemBERTa architecture — ChemBERTa-77M-MLM and ChemBERTa-77M-MTR—are evaluated to assess the impact of pretraining objective alignment on downstream $pIC_{50}$ regression. Both are RoBERTa-based transformers with approximately 77 million parameters, pretrained on 77 million canonical SMILES strings, but differ fundamentally in their self-supervised tasks. ChemBERTa-MLM employs standard Masked Language Modeling (MLM), wherein the model learns to reconstruct randomly masked tokens in SMILES sequences, thereby acquiring knowledge of syntactic and topological molecular structure without explicit property supervision. In contrast, ChemBERTa-MTR utilizes Masked Token Regression (MTR), a multi-task pretraining objective that simultaneously reconstructs masked tokens and predicts continuous molecular properties associated with the masked substructures; this explicitly aligns the pretraining task with downstream regression. For both variants, three training configurations are assessed: (1) standard training with oversampling, (2) sample weighting with default hyperparameters, and (3) sample weighting with hyperparameters optimized via Optuna. All models replace the original classification head with a single-neuron regression head and are fine-tuned using a weighted mean squared error loss

$$\mathcal{L} = \frac{1}{N} \sum_{i=1}^{N} w_i (y_i - \hat{y}_i)^2,$$

where $y_i$ and $\hat{y}_i$ denote the true and predicted pIC$_{50}$ values, respectively, and $w_i$ is the sample weight as defined in Section 4.1. This systematic comparison enables empirical validation of whether task-aligned pretraining (MTR) combined with distribution-preserving training (sample weighting) yields superior performance for continuous bioactivity prediction.

### 4.4 Performance Comparisons Across Methodological Axes

To rigorously establish the optimal fine-tuning protocol for TDP1 pIC$_{50}$ regression, a controlled ablation study across three interdependent methodological dimensions is conducted, ensuring all comparisons are performed on identical data splits and evaluation metrics. The first axis investigates the impact of data handling strategy on model behavior. Specifically, ChemBERTa-MLM and ChemBERTa-MTR are trained under two distinct regimes: (1) stratified train-only oversampling to a 50:50 active–inactive balance, and (2) sample weighting that preserves the original activity distribution while emphasizing rare active compounds through per-sample loss weights. Both configurations employ a fixed set of default hyperparameters.

The second axis examines the contribution of hyperparameter optimization. Here, both ChemBERTa variants are trained exclusively under the sample weighting strategy, contrasting performance between the default hyperparameter configuration and one optimized via Bayesian search. The optimization, performed using Optuna over 20 trials, targets the minimization of validation set Mean Absolute Error (MAE) and explores a search space encompassing learning rate, weight decay, and dropout. Early stopping with a patience of five epochs is used to mitigate overfitting.

The third and final axis constitutes the comprehensive model architecture comparison, which synthesizes insights from the prior ablations. This evaluation includes four models: a random prediction baseline, a Random Forest regressor trained on Morgan fingerprints and RDKit descriptors, and the two ChemBERTa variants—MLM and MTR—both configured with the superior data strategy (sample weighting) and the refined, Optuna-optimized hyperparameters. This unified setup isolates the effect of the pretraining objective itself, providing a fair assessment of whether the property-aware MTR pretraining yields a tangible advantage over the structure-focused MLM for the continuous pIC$_{50}$ prediction task.

### 4.5 Synthesis of Training Strategies and Final Model Selection

The results across these three comparative axes converge to a coherent and empirically grounded selection of the final model. The data strategy comparison demonstrates that sample weighting consistently outperforms oversampling in terms of early enrichment metrics, such as Enrichment Factor at 1% (EF$_1$%) and Precision@1%, for both ChemBERTa variants. This superiority is attributed to the preservation of the true pIC$_{50}$ distribution, which enables the model to generalize more effectively to the high-potency region of the activity landscape.

The hyperparameter optimization study further reveals that a tailored configuration—featuring a smaller batch size, reduced weight decay, and higher dropout—yields a measurable improvement in validation loss and, crucially, in virtual screening utility. Both ChemBERTa-MLM and

ChemBERTa-MTR benefit from this optimization, confirming that default hyperparameters, while functional, are suboptimal for this specific task and data regime.

Finally, the architecture comparison establishes that ChemBERTa-MTR, when equipped with the optimal data strategy and hyperparameters, achieves the highest performance among all evaluated deep learning models. Its pretraining objective, which explicitly regresses molecular properties during the masked token reconstruction phase, provides a distinct advantage for the continuous $pIC_{50}$ prediction task over the purely structural focus of MLM.

By synthesizing these findings, we select ChemBERTa-MTR with sample weighting and Optuna-optimized hyperparameters as our final fine-tuned model. This selection is not based on a single metric but on a holistic assessment of regression accuracy, ranking capability, and, most importantly, practical utility for virtual screening, as quantified by $EF_1\%$ and Precision@K. This systematic validation ensures that our final pipeline is not merely a heuristic choice but a data-driven optimum.

# 5 Experiments

This section presents a systematic experimental evaluation of fine-tuned ChemBERTa models for the quantitative prediction of TDP1 inhibitory activity. The evaluation framework is structured to address four central questions: first, how transformer-based architectures compare against established classical machine learning baselines; second, whether the choice of pretraining objective—Masked Language Modeling versus Masked Token Regression—confers a measurable advantage for continuous $pIC_{50}$ regression; third, how the selection of data handling strategy, specifically oversampling against sample weighting, influences model behavior on the critical subset of high-potency compounds; and fourth, what the ultimate utility of the optimized model is in a realistic virtual screening scenario. To ensure valid and actionable conclusions, all models are trained and evaluated on identical stratified data splits, and performance is assessed through a hierarchy of metrics that prioritize practical drug discovery value over purely statistical fit.

## 5.1 Evaluation Metrics: Definitions and Rationale

Given the extreme class imbalance inherent in high-throughput screening data—where active compounds constitute only 2.1% of the dataset—conventional regression metrics such as mean squared error, mean absolute error, and the coefficient of determination are of limited utility. These metrics are overwhelmingly dominated by the prediction accuracy on the abundant inactive compounds and fail to reflect a model's capacity to identify the rare, high-value actives that drive lead optimization.

Consequently, our evaluation prioritizes metrics that directly quantify screening efficiency and practical utility. The $EF_x\%$ measures the fold-increase in the concentration of active compounds within the top x% of the model-ranked list relative to random selection. An $EF_1\%$ of 15.3, for instance, signifies that the model enriches actives in the top 1% at a rate 15.3 times greater than chance, a critical indicator of cost-saving potential in experimental validation. Complementing this, Precision@K reports the absolute fraction of active compounds found among the top K predictions.

While these early-recognition metrics are paramount, the Area Under the Receiver Operating Characteristic Curve (AUC-ROC) serves as a robust, threshold-independent measure of the model's overall discrimination capability. It evaluates the model's performance across all possible classification thresholds, offering a holistic view of its ranking quality that contextualizes the more focused enrichment results. Together, these three metrics—Enrichment Factor, Precision@K, and AUC-ROC—form a coherent and application-grounded evaluation protocol that aligns our computational results with the practical imperatives of modern drug discovery, complemented by the Ranking Differential, which quantifies the mean separation between predicted $pIC_{50}$ values of active and inactive compounds as a direct measure of discriminatory power.

## 5.2 Ablation Study: Data Strategy (Oversampling vs. Sample Weighting)

To assess the influence of data handling on model performance, oversampling and sample weighting is compared using ChemBERTa-MLM and ChemBERTa-MTR under identical hyperparameters.

As shown in Figure 3, sample weighting consistently outperforms oversampling across the most key virtual screening metrics. For both models, it yields higher $EF_1\%$ and greater Precision@1%, indicating superior early enrichment of active compounds. Although AUC-ROC improves only marginally, the gains in $EF_1\%$—from 11.1× to 17.4× for MTR—demonstrate that sample weighting better supports the practical goal of identifying potent inhibitors with minimal experimental screening. These results confirm that preserving the natural activity distribution during training leads to more effective models for real-world drug discovery.

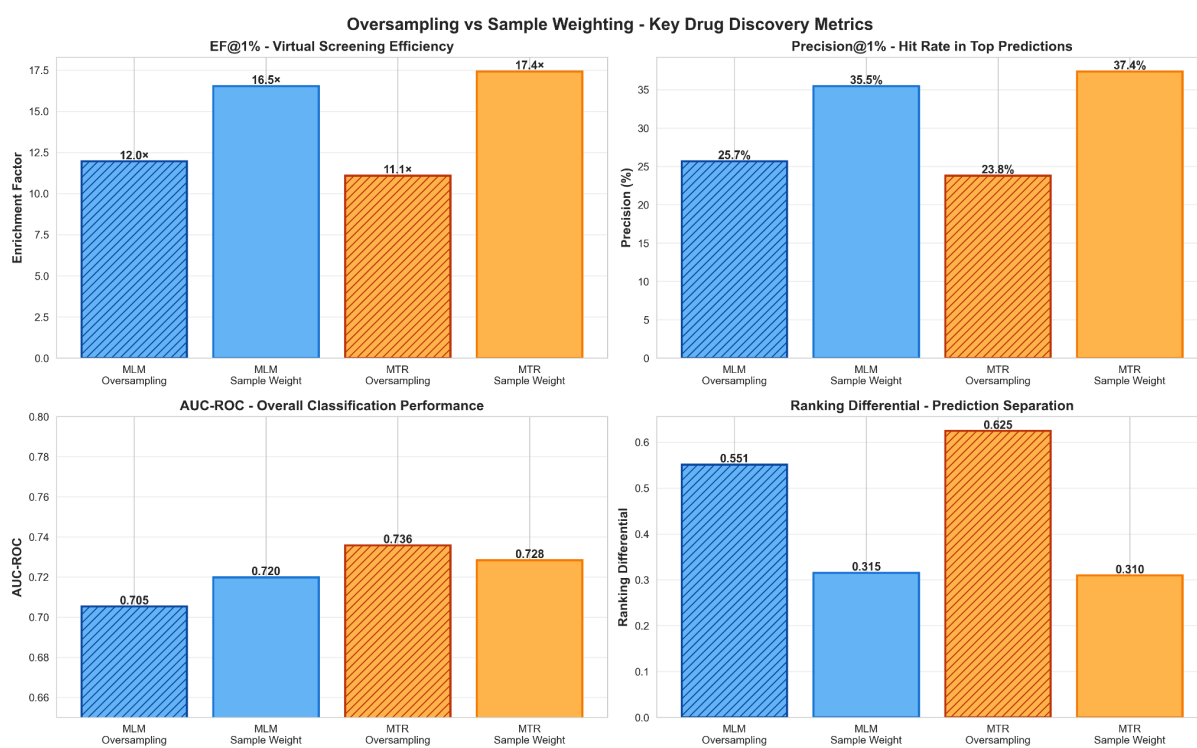

Figure 3: Key Metrics Plots for different Data Strategies

## 5.3 Hyperparameter Optimization with Optuna

To refine the performance of ChemBERTa-MTR under sample weighting, Bayesian hyperparameter optimization is employed using Optuna, targeting the minimization of Active MAE on the validation set. The search space included learning rate, weight decay, and dropout, with twenty trials conducted.

The optimized configuration—learning rate $4.61 \times 10^{-5}$, batch size 32, weight decay 0.0011, dropout 0.100, and 7 epochs—yielded a substantial improvement in virtual screening efficiency. As shown in Figure 4, for both ChemBERTa-MLM and ChemBERTa-MTR, the application of Optuna led to significant gains in $EF_1\%$ and Precision@1%. For MTR, $EF_1\%$ increased from 15.3× to 17.4×, while Precision@1% rose from 32.8% to 37.4%, indicating a higher hit rate among top predictions.

These results demonstrate that hyperparameter optimization is not merely an incremental refinement but a critical step in unlocking the full potential of pre-trained chemical language models for TDP1 inhibitor discovery.

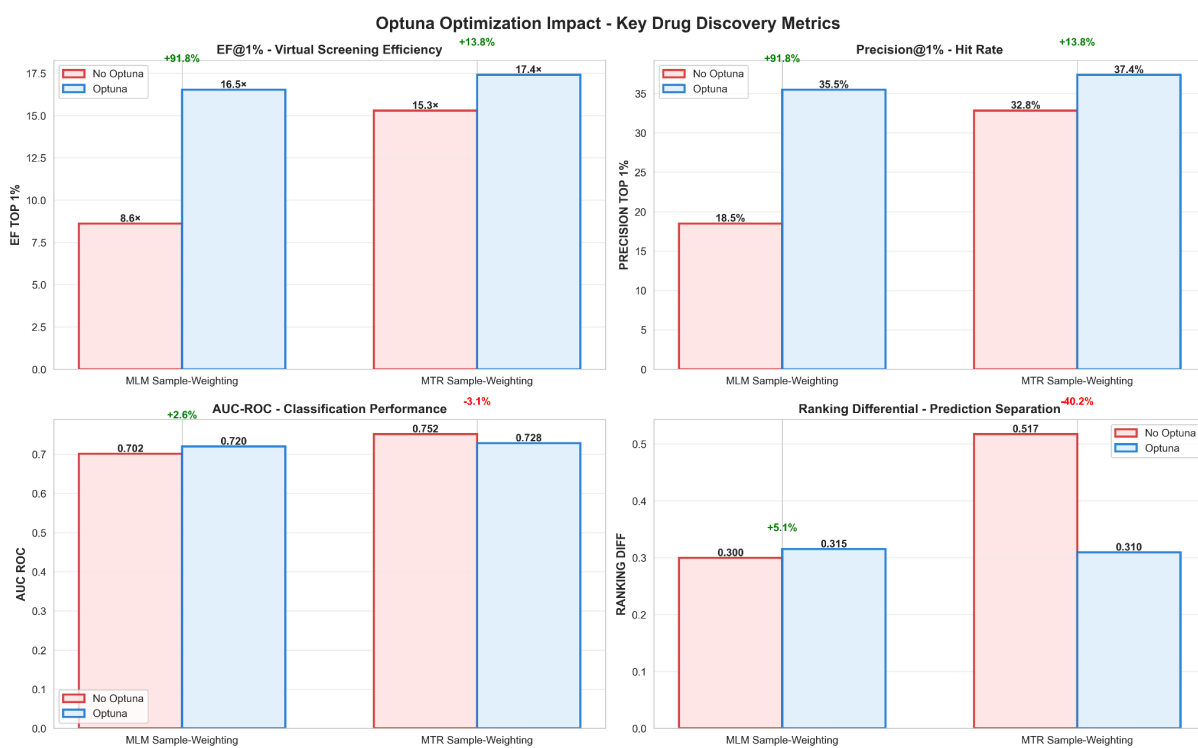

Figure 4: Key Metrics Plots for Hyperparameter Optimization

## 5.4 Baseline Comparison: Classical ML & Transformers

To contextualize the performance of our fine-tuned transformer models, two fundamental baselines is established. The first is a random predictor, which generates $pIC_{50}$ values by uniform sampling within the observed range of the training set from 3.85 to 9.10. This serves as a theoretical lower bound: any meaningful model must demonstrably outperform this null hypothesis. The second baseline is a Random Forest regressor trained on 2048-bit Morgan fingerprints with radius = 2, a well-established and highly effective feature representation in cheminformatics.

All models, including the baselines, are evaluated on the identical stratified test set to ensure a fair comparison. As shown in Figure 5, the Random Forest model achieves an $EF_1\%$ of 21.5× and a Precision@1% of 46.0%, significantly outperforming both ChemBERTa variants. In contrast, the random predictor yields EF$_1$% round 1.0× and Precision@1% around 2.1%, confirming its role as a lower-bound reference.

While the transformer-based models do not surpass Random Forest in this study, ChemBERTa-MTR with sample weighting and Optuna-tuned hyperparameters achieves a competitive EF$_1$% of 17.4× and Precision@1% of 37.4%. This demonstrates that end-to-end SMILES-based deep learning can deliver high practical utility for virtual screening, offering a strong alternative to classical methods that rely on hand-crafted features. The results validate our fine-tuning protocol and position ChemBERTa-MTR as a robust, scalable solution for TDP1 inhibitor prioritization.

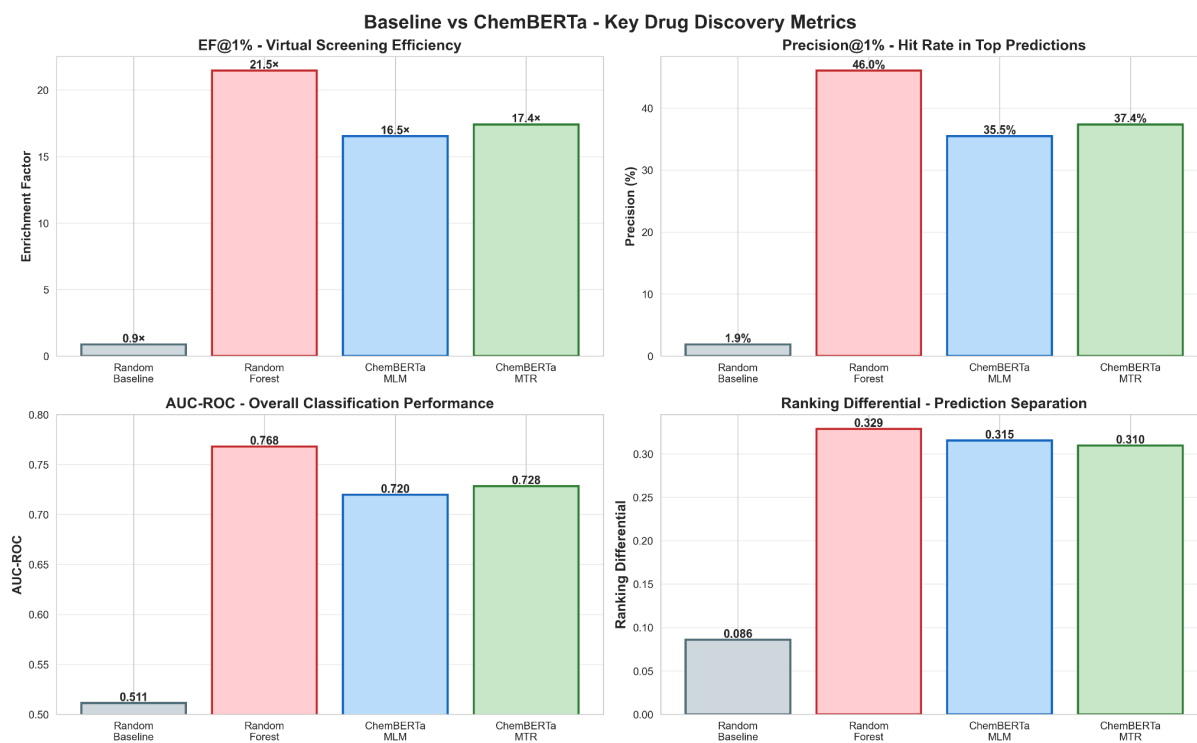

Figure 5: Key Metrics Plots for Baseline Comparison

## 5.5 Final Model Performance and Virtual Screening Utility

The final fine-tuned ChemBERTa-MTR model, trained with sample weighting and optimized hyperparameters, is evaluated on the held-out test set to assess its predictive accuracy and practical utility for virtual screening. The aforementioned results collectively demonstrate that the final model is not only statistically sound but also practically effective. Its strong early enrichment capability—quantified by an Enrichment Factor at 1% of 17.4× and a Precision@1% of 37.4%—confirms its potential to significantly reduce experimental cost and accelerate the discovery of novel TDP1 inhibitors.

## 6 Conclusion

This work presents a systematic evaluation of fine-tuned ChemBERTa models for the quantitative prediction of TDP1 inhibitory activity from SMILES strings. By integrating a large-scale consensus dataset of 177,092 compounds and employing a rigorously controlled experimental design, we demonstrate that the combination of ChemBERTa-MTR, stratified data splitting, and sample weighting yields a strong deep learning pipeline for this task. This approach preserves the true activity distribution, mitigates the challenges of severe class imbalance, and achieves competitive virtual screening performance—most notably an Enrichment Factor at 1% of 17.4×.

## 6.1 Analysis and Limitations

The superior performance of the classical Random Forest baseline, which attained an $EF_1\%$ of 21.5×, can be attributed to several factors. First, the dataset size around 177K compounds lies within the regime where well-engineered molecular features—such as Morgan fingerprints—combined with a robust, interpretable model like Random Forest often outperform more complex architectures. Second, the explicit encoding of substructural motifs in fingerprints provides a direct and efficient signal for TDP1 inhibition that is immediately leveraged by tree-based models.

Despite this, deep learning models such as ChemBERTa retain significant strategic value. They enable end-to-end learning directly from raw SMILES, eliminating the need for manual feature engineering and reducing pipeline complexity. More importantly, their capacity is expected to scale favorably with larger datasets, where the benefits of self-supervised pretraining can be fully realized. Finally, ChemBERTa's native SMILES-based representation provides a natural foundation for future integration with generative models, enabling the direct design of novel inhibitor structures—an avenue not readily accessible with fingerprint-based approaches. Thus, while Random Forest serves as a strong baseline, ChemBERTa represents a more extensible and forward-looking paradigm for AI-driven drug discovery.

## 6.2 Summary

This study presents a systematic evaluation of fine-tuned ChemBERTa models for the quantitative prediction of TDP1 inhibitory activity $pIC_{50}$ using a large, imbalanced dataset of 177,092 compounds. By integrating data from multiple public sources—including three PubChem bioassays and the Goethe University consensus dataset—a robust benchmark is established for regression-based virtual screening in the context of DNA repair–targeted oncology.

Our experimental framework rigorously compares ChemBERTa-MLM and ChemBERTa-MTR across two data handling strategies—oversampling and sample weighting—and benchmarks them against a random predictor and a Random Forest model trained on Morgan fingerprints. The results demonstrate that sample weighting is markedly superior to oversampling for continuous $pIC_{50}$ prediction. By preserving the natural activity distribution, sample weighting avoids the distortion of the $pIC_{50}$ landscape inherent in oversampling, thereby enabling more effective generalization to high-potency compounds and significantly improving early enrichment metrics such as $EF_1\%$ and Precision@1%.

The final optimized model, ChemBERTa-MTR with sample weighting and Optuna-tuned hyperparameters, achieves an $EF_1\%$ of 17.4× and Precision@1% of 37.4%, establishing it as a practical and deployable tool for prioritizing candidates in experimental screening campaigns. The pipeline operates end-to-end from canonical SMILES input to $pIC_{50}$ output, requires no 3D conformational data, and is fully reproducible using open-source tools.

For future work, we propose three directions: (1) incorporating uncertainty quantification to assess prediction reliability; (2) extending the framework to multi-target or polypharmacology prediction using the broad target coverage of the consensus dataset; and (3) integrating the model with generative chemistry approaches to enable de novo design of TDP1 inhibitors.

In summary, this work demonstrates that chemical language models, when aligned with task-appropriate data strategies and evaluated through a drug discovery–oriented lens, can transform large-scale, noisy bioactivity data into a strategic asset for accelerating therapeutic innovation.

## Acknowledgements

The author acknowledges the use of AI models including Grok, Qwen and Claude with any versions to assist with language polishing and grammatical refinement of this report. All scientific content, experimental design, data analysis, interpretation, and conclusions are the sole work of the author. The AI tool was used exclusively for improving clarity and fluency of the text and did not contribute to any aspect of the research or intellectual content.